\documentclass[review]{elsarticle}
\graphicspath{ {./figures/} }
\usepackage{hyperref}
\usepackage{amsmath}
\usepackage{subfigure}
\usepackage{float}
\usepackage{verbatim} 
\usepackage{apalike}
\usepackage[capitalize]{cleveref}
\restylefloat{figure}
\restylefloat{table}

\biboptions{authoryear}

\begin{document}
\begin{frontmatter}

\title{Obstacle Avoidance for UAS in Continuous Action Space Using Deep Reinforcement Learning}

\author[label1]{Jueming Hu}

\author[label2]{Xuxi Yang}

\author[label3]{Weichang Wang}

\author[label4]{Peng Wei}

\author[label5]{Lei Ying}

\author[label1]{Yongming Liu \corref{cor1}}

\cortext[cor1]{Corresponding author.}
\address[label1]{School for Engineering of Matter, Transport \& Energy, Arizona State University, Tempe, AZ, US}
\address[label2]{Department of Aerospace Engineering, Iowa State University, Ames, IA, US}
\address[label3]{Computer \& Energy Engineering, Arizona State University, Tempe, AZ, US}
\address[label4]{Department of Mechanical and Aerospace Engineering, George Washington University, Washington, DC, US}
\address[label5]{Engineering and Computer Science Department,University of Michigan, Ann Arbor, Ann Arbor, MI, US}

\begin{abstract}
Obstacle avoidance for small unmanned aircraft is vital for the safety of future urban air mobility (UAM) and Unmanned Aircraft System (UAS) Traffic Management (UTM). There are many techniques for real-time robust drone guidance, but many of them solve in discretized airspace and control, which would require an additional path smoothing step to provide flexible commands for UAS. To provide a safe and efficient computational guidance of operations for unmanned aircraft, we explore the use of a deep reinforcement learning algorithm based on Proximal Policy Optimization (PPO) to guide autonomous UAS to their destinations while avoiding obstacles through continuous control. The proposed scenario state representation and reward function can map the continuous state space to continuous control for both heading angle and speed. To verify the performance of the proposed learning framework, we conducted numerical experiments with static and moving obstacles. Uncertainties associated with the environments and safety operation bounds are investigated in detail. Results show that the proposed model can provide accurate and robust guidance and resolve conflict with a success rate of over 99\%.
\end{abstract}


\end{frontmatter}

\section{Introduction}
From delivery drones to autonomous electrical vertical take-off and landing (eVTOL) passenger aircraft, modern unmanned aircraft systems (UAS) can perform many different tasks efficiently, including goods delivery, surveillance, public safety, weather monitoring, disaster relief, search and rescue, traffic monitoring, videography, and air transportation \citep{balakrishnan2018blueprint,kopardekar2016unmanned}. Urban air mobility (UAM) is likely to occur in urban areas close to buildings or airports. Thus, it is expected that UAS can use onboard detect and avoid systems to avoid other traffic, hazardous weather, terrain, and man-made and natural obstacles without constant human intervention \citep{kopardekar2016unmanned}.

Many mathematical models for aircraft conflict resolution have been
proposed in the literature. Research efforts can be divided into centralized algorithms and decentralized algorithms. Centralized methods can be based on semidefinite programming \citep{frazzoli2001resolution}, nonlinear programming \citep{raghunathan2004dynamic,enright1992discrete}, mixed-integer linear programming \citep{schouwenaars2001mixed,richards2002aircraft,pallottino2002conflict,vela2009mixed}, mixed-integer quadratic programming \citep{mellinger2012mixed}, sequential convex programming \citep{augugliaro2012generation,morgan2014model}, second-order cone programming \citep{acikmese2007convex}, evolutionary techniques \citep{delahaye2010aircraft,cobano2011path}, and particle swarm optimization \citep{pontani2010particle}. These centralized methods often pursue the global optimum for all the aircraft. However, as the number of aircraft grows, the computation cost of these methods typically scales exponentially. Among the decentralized methods, the conflict resolution problem can also be formulated as a Markov Decision Process (MDP). Reinforcement
Learning (RL) has been proved to be a good solution to aircraft traffic management, but mostly use traditional algorithms \citep{sun2019reinforcement}. The next-generation airborne collision avoidance system (ACAS X) formulates the collision avoidance systems (CAS) problem as a partially observable Markov Decision Process (POMDP) and has been extended to unmanned aircraft, named ACAS Xu \citep{kochenderfer2012next}. Both ACAS X and ACAS Xu use Dynamic Programming (DP) to determine the expected cost of each action \citep{manfredi2016introduction, owen2019acas}. \cite{chryssanthacopoulos2012decomposition} combined decomposition methods and DP for optimized collision avoidance with multiple threats. The traditional RL algorithms require a fine discretization scheme of state space and finite action space. Discretization potentially reduces safety by adding discretization errors and cannot provide flexible maneuver guidance for UAS. In addition, discretizing large airspace implies a high computation demand and can be time-consuming.
Tree search based algorithms \citep{yang2018autonomous,yang2020scalable} are also applied to CAS problems using MDP formulation which does not involve state discretization. But they typically require high onboard computation time to accommodate the continuous state space.

The large and continuous state and action spaces present a challenge for conflict resolution problems using reinforcement learning. Recently, Deep Reinforcement Learning (DRL) is studied to solve this challenge by applying the deep neural network to approximate the cost and the optimal policy functions. Development of DRL algorithms, such as Policy Gradient \citep{sutton2000policy}, Deep Q-Networks (DQN) \citep{mnih2013playing}, Double DQN \citep{van2016deep}, Dueling DQN \citep{wang2015dueling}, Deep Deterministic Policy Gradient (DDPG) \citep{lillicrap2015continuous}, Asynchronous Advantage Actor-Critic (A3C) \citep{mnih2016asynchronous}, and Proximal Policy Optimization (PPO) \citep{ppo2} has increased the potential of automation. \cite{li2019optimizing} used DQN to compute corrections for an existing collision avoidance approach to account for dense airspace. In \cite{yang2019real}, the feasibility of using algorithms based on DQN in UAV obstacle avoidance is verified. \cite{wulfeuav} concluded that DQN can outperform value iteration both in terms of evaluation performance and solution speed when solving a UAV collision avoidance problem. The performance of the agent in avoiding single up to multiple aircraft by using the DQN algorithm is investigated in \cite{keong2019reinforcement}. \cite{brittain2020deep} proposed a novel deep multi-agent reinforcement learning framework based on PPO to identify and resolve conflicts among a variable number of aircraft in a high-density, stochastic, and dynamic sector in en-route airspace. The DRL work mentioned above is in continuous state and discrete action space.

There has been less progress on utilizing DRL to solve UAS conflict resolution with continuous control. \cite{pham2019machine} proposed a method inspired by Deep Q-learning and Deep Deterministic Policy Gradient algorithms and it can resolve conflicts, with a success rate of over 81 \%, in the presence of traffic and varying degrees of uncertainties. \cite{ma2018saliency} developed a generic framework that integrates an autonomous obstacle detection module and an actor-critic based reinforcement learning (RL) module to develop reactive obstacle avoidance behavior for a UAV. Experiments in \cite{ppo2} test PPO on a collection of benchmark tasks, including simulated robotic locomotion and Atari game playing, and show that PPO outperforms other online policy gradient methods. PPO appears to be a favorable balance between sample complexity, simplicity, and wall-time. Thus, a PPO-based conflict resolution model is very valuable for UAS traffic management, which is the major motivation of this study.

To the best of the authors' knowledge, this is the first study to develop a DRL approach based on PPO algorithm to allow the UAS to navigate successfully in continuous state and action spaces. The benefit of calculating in continuous space is that there is no need to discretize the state space or smooth results for postprocessing results. The proposed model with the optimal policy after offline training can be utilized for UAS real-time online trajectory planning.
The main contributions of this paper are as
follows:
\begin{itemize}
    \setlength\itemsep{0em}
    \item  A PPO-based framework has been proposed for UAS to avoid both static and moving obstacles in continuous state and action spaces.
    \item  A novel scenario state representation and reward function are developed and can effectively map the environment to maneuvers. The trained model can generate continuous heading angle commands and speed commands.
    \item  We have tested the effectiveness of the proposed learning framework in the environment with static obstacles, the environment with static obstacles and UAS position uncertainty, and the deterministic and stochastic environments with moving obstacles. Results show that the proposed model can provide accurate and robust guidance and resolve conflict with a success rate of over 99\%.
\end{itemize}

The remainder of this paper is organized as follows. \cref{background} describes the backgrounds of Markov Decision Process and Deep Reinforcement Learning. \cref{mdp} presents the model formulation using Markov Decision Process for UAS conflict resolution in continuous action space. In \cref{result}, the numerical experiments are presented to show the capability of the proposed approach to make the UAS learn to avoid conflict. \cref{Conclusion} concludes this paper.

\section{Background}
\label{background}
In this section, we briefly review the backgrounds of Markov Decision Process (MDP) and Deep Reinforcement Learning (DRL).

\subsection{Markov Decision Process (MDP)}
Since the 1950s, MDPs \citep{Bel} have been well studied and applied to a wide area of disciplines \citep{howard1964dynamic,white1993survey,feinberg2012handbook}, including robotics \citep{koenig1998xavier,thrun2002probabilistic}, automatic control \citep{mariton1990jump}, economics, and manufacturing. In an MDP, the agent may choose any action $a$ that is available based on current state $s$ at each time step. The process responds at the next time step by moving into a new state $s'$ with certain transition probability and gives the agent a corresponding reward $r$. 

More precisely, the MDP includes the following components:
\begin{enumerate}
  \item The state space $\mathcal{S}$ which consists of all the possible states.
  \item The action space $\mathcal{A}$ which consists of all the actions that the agent can take.
  \item Transition function $\mathcal{T}(s_{t+1}|s_t,a_t)$ which describes the probability of arriving at state $s_{t+1}$, given the current state $s_t$ and action $a_t$.
  \item The reward function $r(s_t,a_t,s_{t+1})$ which decides the immediate reward (or expected immediate reward) received after transitioning from state $s$ to state $s'$, due to action $a$. In general, the reward will depend on the current state, current action, and the next state. $r_t$ is the immediate reward at the time step $t$ and $R_t$ is the total discounted reward from time-step $t$ forwards.
  \item A discount factor $\gamma \in [0,1]$ which decides the preference for immediate reward versus future rewards. Setting the discount factor less than 1 is also beneficial for the convergence of cumulative reward.
\end{enumerate}

In an MDP problem, a policy $\pi$ is a mapping from the state to 
a distribution over actions (known as stochastic policy)
\begin{equation*}
\pi: \mathcal{S} \rightarrow \text{Prob}(\mathcal{A})
\end{equation*}
or to 
one specific action (known as deterministic policy) 
\begin{equation}
\pi:\mathcal{S} \rightarrow \mathcal{A}
\end{equation}
 
The goal of MDP is to find an optimal policy $\pi^*$ that, if followed from any initial state, maximizes the expected cumulative immediate rewards:
\begin{equation}
    \begin{aligned}
        \pi^* &= \underset{\pi}{\arg\max} E [R_t|\pi]\\
        &= \underset{\pi}{\arg\max} E[\sum_{t'=t}^T \gamma^{t'-t} r(s_{t'},a_{t'})|\pi]
    \end{aligned}
\end{equation}

Q-function and value function are two important concepts in MDP. 
The optimal Q-function $Q^*(s, a)$ represents the expected cumulative reward received by an agent that  starts from state $s$, picks action $a$, and chooses action optimally afterward. Therefore, $Q^*(s, a)$ is an indication of how good it is for an agent to pick action $a$ while being at state $s$.
The optimal value function $V^  *(s)$ denotes the maximum expected cumulative reward when starting from state $s$, which can be expressed as the maximum of $Q^*(s, a)$ over all possible actions:
\begin{equation}
  V^{*}(s)=\max _{a} Q^{*}(s, a), \quad \forall s \in \mathcal{S}
\end{equation}

\subsection{Deep Reinforcement Learning}
Reinforcement learning \citep{sutton2018reinforcement} is an efficient algorithm to solve the MDP problem. With the advent of deep learning, Deep Reinforcement Learning (DRL) achieved much success recently, including game of GO \citep{silver2017mastering}, Atari games \citep{mnih2015human,hessel2018rainbow}, Warcraft \citep{vinyals2017starcraft}. In general, deep reinforcement learning can be divided into value-based learning \citep{mnih2015human,lillicrap2015continuous} and policy-based algorithm \citep{mnih2016asynchronous,schulman2015trust,ppo2}. In this paper, we consider a policy-based DRL algorithm to generate policies for the agent. Comparing with value-based DRL algorithms, the policy-based algorithm is effective in high-dimensional or continuous action spaces and can learn stochastic policies, which is beneficial when there is uncertainty in the environment.

Typically, the policy-based algorithm uses function approximation such as neural networks to approximate the policy $\pi(s)$, where the input is the current state and output is the probability of each action (for discrete action space) or an action distribution (for continuous action space). After each trajectory $\tau$, the algorithm updates the parameter of the function approximation to maximize the cumulative reward using gradient ascent:
\begin{equation}
\label{pg}
    \nabla_{\theta} J\left(\pi_{\theta}\right)=\underset{\tau \sim \pi_{\theta}}{\mathrm{E}}\left[\sum_{t=0}^{T} \nabla_{\theta} \log \pi_{\theta}\left(a_{t} | s_{t}\right) R_t\right]
\end{equation}
where $J(\pi_\theta)$ is the expected cumulative reward of policy $\pi$ parameterized by $\theta$, $\pi_\theta(a_t|s_t)$ is the probability of action $a_t$ for state $s_t$, and $R_t$ is the cumulative reward gathered by the agent for the remaining trajectory $\tau$ in one episode. The general idea of \cref{pg} is to reduce the probability of sampling an action that leads to a lower return and increase the probability of action leads to a higher reward. But one issue is that the cumulative reward usually has very high variance, which makes the convergence speed to be slow. To address this issue, researchers proposes actor-critic algorithm \citep{sutton2018reinforcement} where a critic function is introduced to approximate the state value function $V(s_t)$. By subtracting the value function $V(s_t)$, the expectation of the gradient keeps unchanged and the variance is reduced dramatically:
\begin{equation} \label{eqn:ac}
    \nabla_{\theta} J\left(\pi_{\theta}\right)=\underset{\tau \sim \pi_{\theta}}{\mathrm{E}}\left[\sum_{t=0}^{T} \nabla_{\theta} \log \pi_{\theta}\left(a_{t} | s_{t} \right) (R_t - V(s_t))\right]
\end{equation}
where the function approximator $V(s_t)$ is updated to approximate the value function for $s_t$.

\section{Markov Decision Process Formulation}
\label{mdp}
In this study, the UAS and intruders are considered as a point mass. The objective of the proposed conflict resolution algorithm is to find the shortest path for a UAS to its goal while avoiding conflict with other UAS and static obstacles. Guiding the UAS to its destination is a discrete-time stochastic control process that can be formulated as a Markov Decision Process (MDP). In the following subsections, we introduce the MDP formulation by describing its state representation, action space, terminal state, and reward function. For this work, a method of deep reinforcement learning, proximal policy optimization algorithm developed in \cite{ppo2}, is adopted. The reason and details are also introduced in this Section.

\subsection{State representation}
The agent gains knowledge of the environment from the state of the formulated MDP. The state should include all the necessary information for an agent to make optimal actions. In this paper, we let $s_t$ denote the agent's state at time $t$. All the parameters are normalized when querying the neural network. 

The state can be divided into two parts, that is $s_t=[s_t^0, s_t^1]$, where $s_t^0$ denotes the part that is related to the agent itself and the goal, and $s_t^1$ denotes the part related to the environment such as obstacles. We use $w_t$ to denote the information of environment (which represents moving/static obstacles in this paper) and set $s_t^1=[w_t^1, w_t^2, ..., w_t^n]$. $w_t^i$ indicates the information of obstacle $i$. To speed up the training of the DRL algorithm, we transform the state by following the robot-centric parameterization in \cite{chen2017decentralized}, where the agent is located at the origin and the x-axis is pointing toward its goal. 

\subsubsection{State representation for static obstacle avoidance}
In the simulations of static obstacle avoidance, 
\begin{equation}
    s_t^0=[d_g, v_x, v_y]
\end{equation}
where $d_g$ is the agent's distance to goal, $v_x, v_y$ denote the agent's velocity. The information of obstacle $i$ is,
\begin{equation}
    w_t^i=[P_y^i, d_i]
\end{equation}
where $P_y^i$ is the position in y-axis of obstacle $i$ and $d_i$ is the agent's distance to the center of obstacle $i$. $P_y^i$ is introduced to help the agent learn the global optimal solution, for example, when approaching the obstacle, turn a small angle counterclockwise if $P_y^i$ is positive, which means the agent is on the right side of the line passing the obstacle center and the goal. We note that $v$ and $P$ are vectors in the transformed coordinate system.

\subsubsection{State representation for moving obstacle avoidance}
As for moving obstacle avoidance, the position of the goal, $(g_x, g_y)$, is added to $s_t^0$,
\begin{equation}
    s_t^0=[d_g, v_x, v_y, g_x, g_y]
\end{equation}
The information of intruder $i$, $w_t^i$ is represented by,
\begin{equation}
    w_t^i=[P_x^i,P_y^i, V_x^i, V_y^i, d_i, V_{{ref}_x}^i, V_{{ref}_y}^i]
\end{equation}
where $P^i$ is the position of intruder $i$, $V^i$ is the velocity of intruder $i$, $d_i$ is the distance between the agent and the intruder $i$ and $V_{ref}^i$ is the velocity of the agent relative to the intruder $i$. We note that $v$, $g$, $P$ and $V$ are vectors in the transformed coordinate system.

\subsection{Action space}

\subsubsection{Action space for static obstacle avoidance and stochastic intruder avoidance}
In the implementations of static obstacle avoidance and stochastic intruder avoidance, the action represents the change in the heading angle for the controlled UAS at each time step. The action space is set to,
\begin{equation}
  \mathcal{A}_h = [-30^\circ, 30^\circ]
\end{equation}
More specifically, at each time step, the agent will select an action $a_h \in \mathcal{A}_h$, and change its heading angle $\psi$:
\begin{equation}
  \psi_{t+1} = \psi_t + a_h
\end{equation}

\subsubsection{Action space for deterministic intruder avoidance}
As for deterministic intruder avoidance case, besides the heading angle change $a_h \in \mathcal{A}_h$, the UAS is also controlled by speed command, which is
updated every one second. During the interval, the two commands are fixed.
Since UAS is more flexible than manned aircraft and there is no available regularization on UAS speed change, UAS speed command $a_v$ can be chosen from $\mathcal{A}_v$,
\begin{equation}
  \mathcal{A}_v = [0 m/s, 40 m/s]
\end{equation}
More specifically, the agent will select an action $a_v \in \mathcal{A}_v$, and change its speed $\mathbf{v}$ at next time step $t+1$:
\begin{equation}
  \mathbf{v}_{t+1} = a_v 
\end{equation}
In real-world applications, however, making a sharp turn is usually not desirable for the controlling of a UAS. Thus a penalty of large heading or speed change due to the power consumption may be considered in future work.

\subsection{Terminal state}
In the current study, the conflict is defined to be when the distance from the agent to the obstacle is less than a minimum separation distance. When the UAS operation is deterministic, a buffer zone is not necessary and the minimum separation distance is set to zero. In the implementations of static obstacle avoidance with uncertainty and moving obstacle avoidance, the UAS position uncertainty is taken into account. The separation requirement is determined according to the operational safety bound proposed in \cite{hu2020probabilistic}. With the UAS speed of 20 $m\slash s$ and other UAS operation performance following the mean value shown in Table 3 in \cite{hu2020probabilistic}, the minimum separation distances for static obstacle avoidance is 75 $m$ and 150 $m$ for moving obstacle avoidance. 

\subsubsection{Terminal state for static obstacle avoidance}
The terminal state for static obstacle avoidance includes two different types of states:

\begin{itemize}
    \item Conflict state: the distance between the agent and obstacle is less than the minimum separation distance.
    \item Goal state: the agent is within 400 $m$ from the destination.
\end{itemize}

\subsubsection{Terminal state for moving obstacle avoidance}
The episode terminates only when the agent is within 200 $m$ from the destination, which indicates the agent accomplishes the navigation task.

\subsection{Reward function}
To guide the agent to reach its goal and avoid conflict, the reward function is developed to award accomplishments while penalizing conflicts or not moving towards the goal.

\subsubsection{Reward function for static obstacle avoidance}
In the simulations of static obstacle avoidance, the reward function, $R(s, a)$, is expressed as the following form, where we set a reward for the goal state and a penalty for the conflict state. The linear term of the reward function guides the UAS flying towards the destination. The constant penalty at each step emphasizes the shortest path rule.
\begin{equation}
\label{reward_single_static}
R(s,a) = 
  \begin{cases}
    10, & \mbox{if $s$ is goal state,}  \\
    -0.001 d_g - 0.05 - 16, & \mbox{if $s$ is conflict state,} \\
    -0.001 d_g - 0.05, & \mbox{otherwise.}
  \end{cases}
\end{equation}

\subsubsection{Reward function for moving obstacle avoidance}
\begin{figure}[hbt!]
    \centering
    \includegraphics[scale=.5]{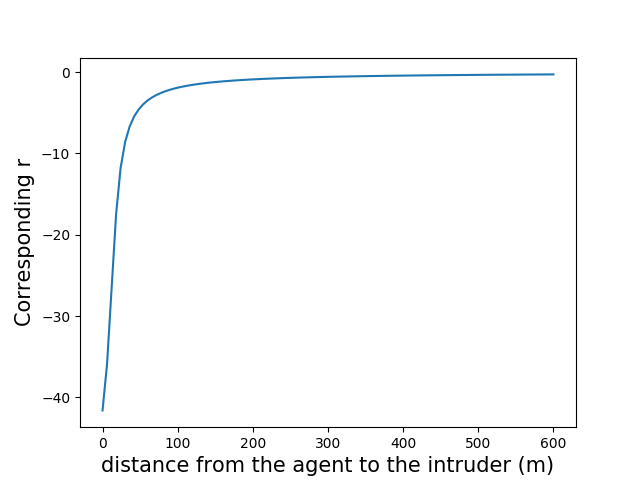}
    \caption{Reward related to the distance from the agent to the intruder.}
    \label{fig:corresponding_r}
\end{figure}

In the simulations of intruder avoidance, the reward function, $R(s, a)$, is expressed in the following form, similar to \cref{reward_single_static}.
\begin{equation}
\label{reward_single_intruder}
R(s,a) = 
  \begin{cases}
    1000, & \mbox{if $s$ is goal state,}  \\
    - c_g d_g - c_0 + \sum_i c_{1i}  (\arctan(c_{2i}(d_i-c_{3i}))-\frac{\pi}{2}) -180 & \mbox{if $s$ is conflict state,} \\
    -c_g d_g - c_0 + \sum_i c_{1i}  (\arctan(c_{2i}(d_i-c_{3i}))-\frac{\pi}{2}) & \mbox{otherwise.}
  \end{cases}
\end{equation}
In this reward function, $c_g, c_0, c_{1i}, c_{2i}, c_{3i}$ are coefficients of different cost and should be balanced to help the agent learn conflict resolution and achieve the goal simultaneously. 
When the ownership is close to the intruder, the inverse tangent term of the reward function is activated to maintain the distance in an appropriate range. With the coefficients set in the stochastic intruder case in \cref{three_case}, the relation between the distance
and the inverse tangent term, $17 (\arctan(0.1 (d_i-12))-\frac{\pi}{2})$, is shown in \cref{fig:corresponding_r}. The agent starts to get a penalty when the distance approaches 250 $m$. This reward setting can help the agent avoid conflicts with other intruders at a relatively early stage. We note that $c_{2i}, c_{3i}$ can be tuned to fit different separation standards.

\subsection{Proximal policy optimization algorithm}
One drawback of $\nabla_\theta J(\pi_\theta)$ proposed in \cite{sutton2018reinforcement}, also shown in \cref{eqn:ac} , is that one bad update can lead to large destructive effects and hinder the final performance of the model. Proximal Policy Optimization (PPO) algorithm \citep{ppo2} was proposed recently to solve this problem by introducing a policy changing ratio describing the change from previous policy to the new policy at time step $t$:
\begin{equation}
  r_t(\theta) = \frac{\pi_\theta(a_t|s_t)}{\pi_{\theta_\text{old}}(a_t|s_t)}
\end{equation}
where $\theta_{\text{old}}$ and $\theta$ denote the network weights before and after update.

By restricting policy changing ratio in the range $[1-\epsilon, 1+\epsilon]$ with $\epsilon$ set to 0.2 in this paper, the PPO loss function for the actor and critic network is formulated as follows:
\begin{equation}
\label{actor_loss}
     L_\pi(\theta) = -\underset{\tau \sim \pi_{\theta}}{\mathrm{E}} [\min (r_t(\theta) \cdot A_t, \text{clip}(r_t(\theta), 1-\epsilon, 1+\epsilon) \cdot A_t] - \beta\cdot H(\pi(\cdot |s_{t}))
\end{equation}
\begin{equation}
\label{critic_loss}
    L_{v} = A_t^2 
\end{equation}
\begin{equation}
    A_t = \| R_t - V(s_t) \|
\end{equation}
where $\epsilon$ is a hyperparameter that bounds the policy changing ratio $r_{t}(\theta)$.
In \cref{actor_loss} and \cref{critic_loss}, the advantage function $A_t$ measures whether or not the action is better or worse than the policy's default behavior.
Also, the policy entropy $\beta\cdot H(\pi(\cdot | s_{t}))$ is added to the actor loss function to encourage exploration by discouraging premature convergence to sub-optimal deterministic polices.

In the implementation, we use two layers Multilayer perceptron (MLP) with 64 hidden units each for both actor and critic networks. Tanh function is chosen as the activation function for the hidden layers. 


\section{Numerical experiments} \label{result}
Numerical experiments are presented in this section to evaluate the proposed conflict resolution model in continuous action space. There are two categories of collision avoidance: static obstacle avoidance and moving obstacle avoidance. As for static obstacle avoidance, we investigate the performance on different obstacle shapes and sizes, and uncertainty in UAS operation. We also study the environment with stochastic intruders under control of heading angle, and the environment with deterministic intruders under control of heading angle and speed. In all the simulations, one pixel in the figure represents 10 $m$ in the real world. The implementation of PPO algorithm is conducted by OpenAI Baselines \citep{baselines}. The deep reinforcement learning model for each case is trained for 30 million time steps.

\subsection{Static obstacle avoidance}
\begin{figure}[hbt!]
    \centering
    \subfigure[]{\includegraphics[width=0.42\textwidth]{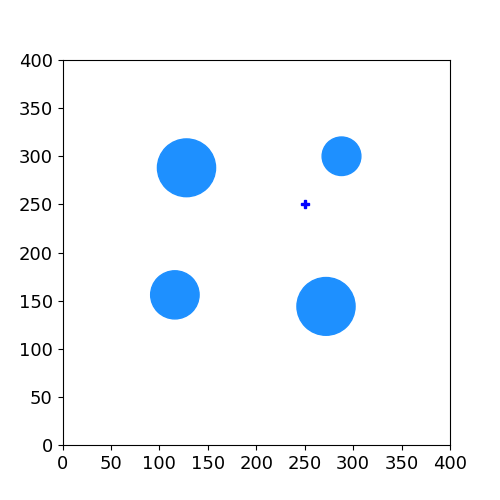}\label{fig:env_cir}} 
    \hspace{1cm}
    \subfigure[]{\includegraphics[width=0.42\textwidth]{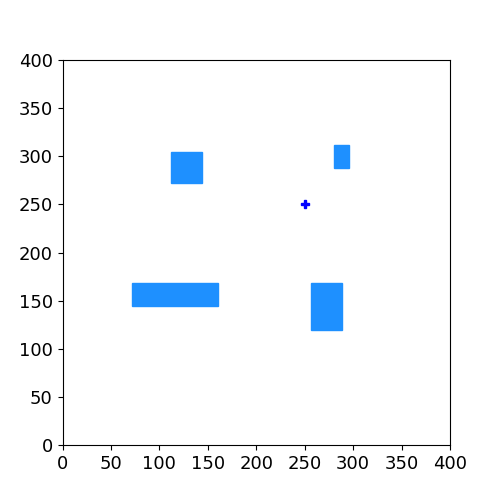}\label{fig:env_rec}} 
    \caption{(axis unit: $\times10$ $m$) (a) Circular obstacle environment.      (b) Rectangular obstacle environment. $+$: goal; blue: no-passing area.}
    \label{fig:env}
\end{figure}

\begin{figure}[hbt!]
    \centering
    \subfigure[]{\includegraphics[width=0.48\textwidth]{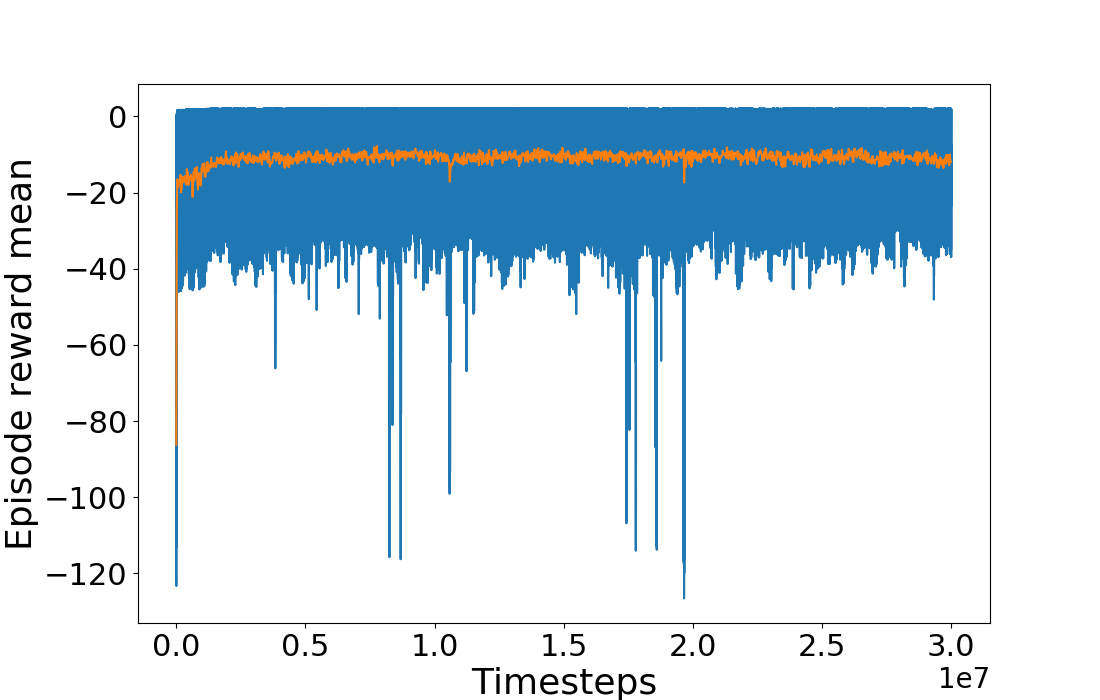}\label{fig:rew_cir}} 
    \subfigure[]{\includegraphics[width=0.48\textwidth]{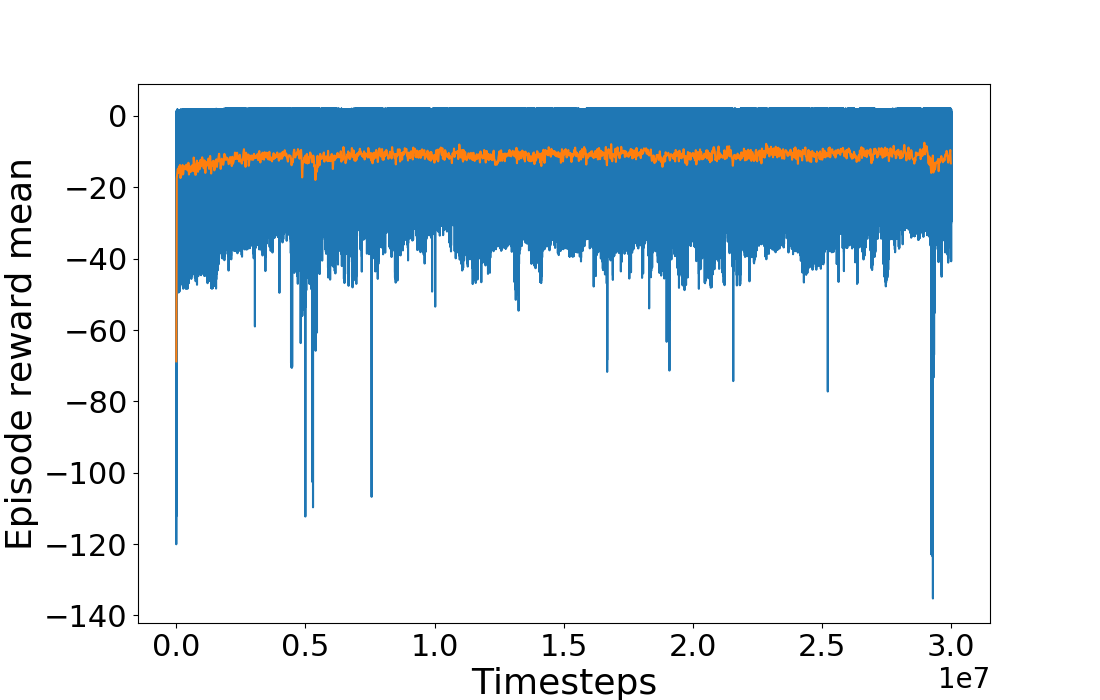}\label{fig:rew_rec}} 
    \subfigure[]{\includegraphics[width=0.48\textwidth]{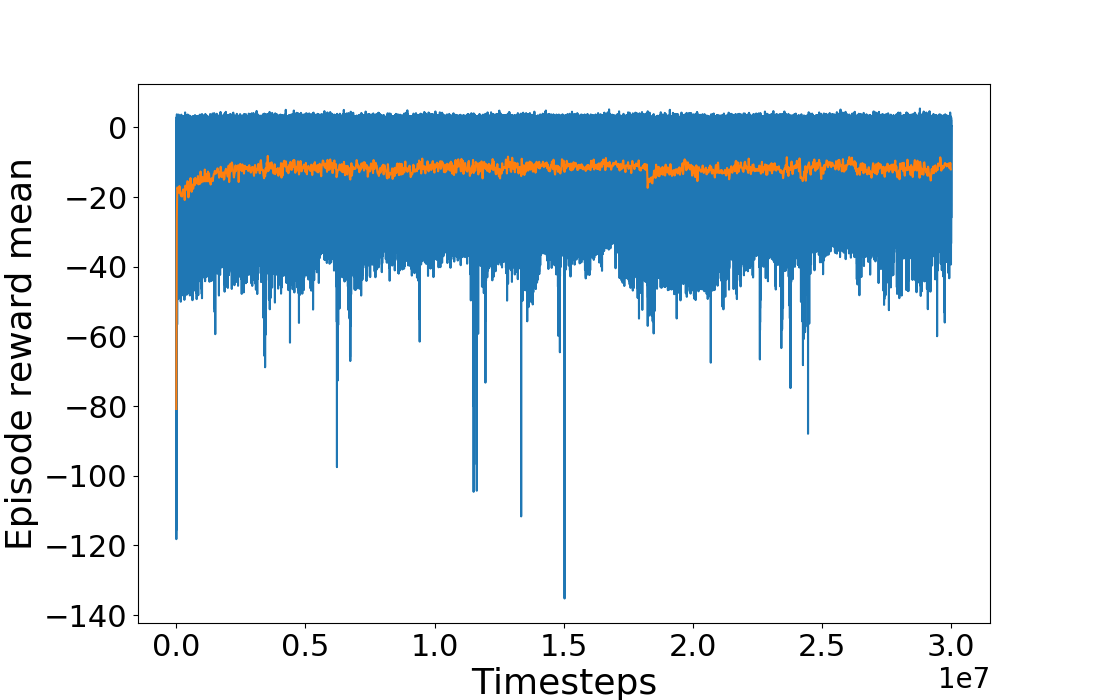}\label{fig:rew_unc}}
    \caption{Episode reward mean for (a) circular obstacle avoidance, (b) rectangular obstacle environment, and (c) circular obstacle avoidance with probabilistic agent's position.}
    \label{fig:rew}
\end{figure}

The simulation environment is the free flight airspace with 4 $km$ length and 4 $km$ width. The UAS speed is set to 20 $m/s$. During the training process, the starting position of the aircraft is randomly sampled from four edges of the airspace boundary and is an array of the integer type for simplification. The goal is located at (2500, 2500). In this experiment, we study two types of static obstacles: circular obstacle and rectangular obstacle, as shown in \cref{fig:env}. The plus sign represents the goal position and the blue region represents the no-passing area. The episode reward mean is shown in \cref{fig:rew}, which shows the episode reward is growing and the policy is converging to the optimal solution. To visualize the performance of the proposed conflict resolution model, we generate a testing set of 160 trajectories starting from different origins.  The origin for testing is chosen every 100 $m$ on each edge of the airspace boundary. Heading angles in 160 trajectories are collected and the heading angle is plotted every 15 time steps.

\subsubsection{Circular obstacle avoidance}
\begin{figure}[hbt!]
    \centering
    \includegraphics[width=0.6\textwidth]{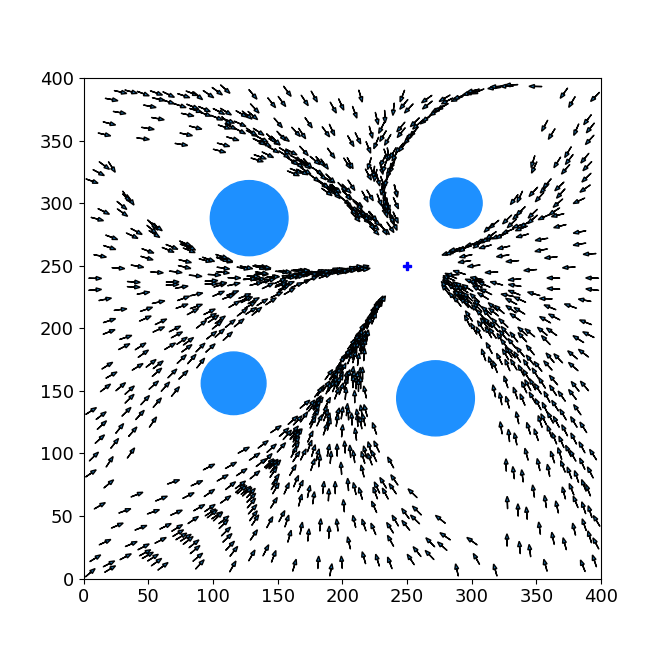}
    \caption{(axis unit: $\times10$ $m$) Results of circular obstacle avoidance. $+$: goal; blue: no-passing area; arrow: the selected heading direction.}
    \label{fig:cir_heading}
\end{figure}

The static obstacle set up for this case study is shown in \cref{fig:env_cir} and the testing result of 160 trajectories starting from different locations on the airspace boundary is shown in \cref{fig:cir_heading}. The black arrow represents the agent's selected heading direction at each position.

From \cref{fig:cir_heading}, it can be seen that the agent is selecting the heading angle pointing to the goal and tending to avoid the no-passing region. Also, the agent chooses the optimal behavior according to the relative position of the agent, obstacle, and goal. For example, near the lower-left obstacle, if the agent's position is above the line passing the obstacle center and the goal, the UAS takes a small left turn to avoid the obstacle. Otherwise, the UAS bypasses the lower semicircle. For the 160 generated trajectories in \cref{fig:cir_heading}, there is no failure.

\subsubsection{Rectangular obstacle avoidance}
\begin{figure}[hbt!]
    \centering
    \includegraphics[width=0.6\textwidth]{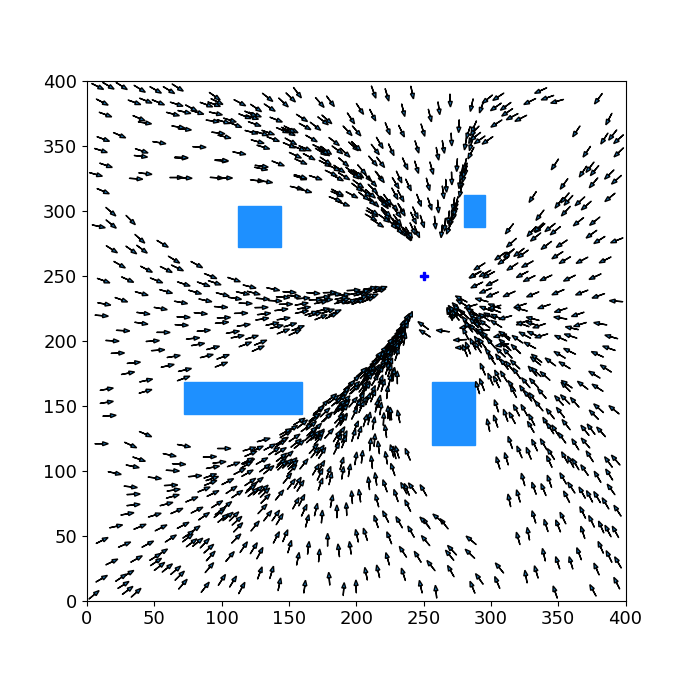}
    \caption{(axis unit: $\times10$ $m$) Results of rectangular obstacle avoidance. $+$: goal; blue: no-passing area; arrow: the selected heading direction.}
    \label{fig:rec_heading}
\end{figure}

The difference between this rectangular obstacle case and the previous circular obstacle case in simulation is the condition when checking whether the agent is at a conflict state. The environment for this case is shown in \cref{fig:env_rec} and the testing result of 160 trajectories is shown in \cref{fig:rec_heading}.

The performance in \cref{fig:rec_heading} is similar to the result in \cref{fig:cir_heading}. The UAS learns to bypass the obstacle through one side of the obstacle depending on the relative position of the agent, obstacle, and goal. No failure happens among the test of 160 trajectories.

Results in \cref{fig:cir_heading} and \cref{fig:rec_heading} show that the proposed model has the capability to make the UAS learn to find the shortest path and also avoid static obstacles for different obstacle sizes or shapes.

\subsubsection{Circular obstacle avoidance with uncertainty}
This case is studied to see the performance of handling uncertainty by the proposed conflict resolution model. UAS operation is stochastic and randomness exists in almost every aspect of UTM. Inclusion of uncertainty quantification of aircraft operation is critical for future safety analysis (e.g., deviation from a trajectory plan due to wind, true speed, positioning error) \citep{hu2020probabilistic, liu2018information, hu2020uas, pang2019recurrent, pang2019aircraft, pang2021data}. Thus, to model the uncertainties in UAS operation, we form a circle, the center of which is the predicted UAS position without uncertainty. And the radius is the separation requirement, 75 $m$. With 90\% probability, the UAS position is accurately located at the center of the circle; with a 10\% probability, the UAS position will be located at a point around the circle with a uniform distribution. Such uncertainty is considered when calculating the agent's position at the next time step after taking action $a$. 

\begin{figure}[H]
    \centering
    \subfigure[]{\includegraphics[width=0.6\textwidth]{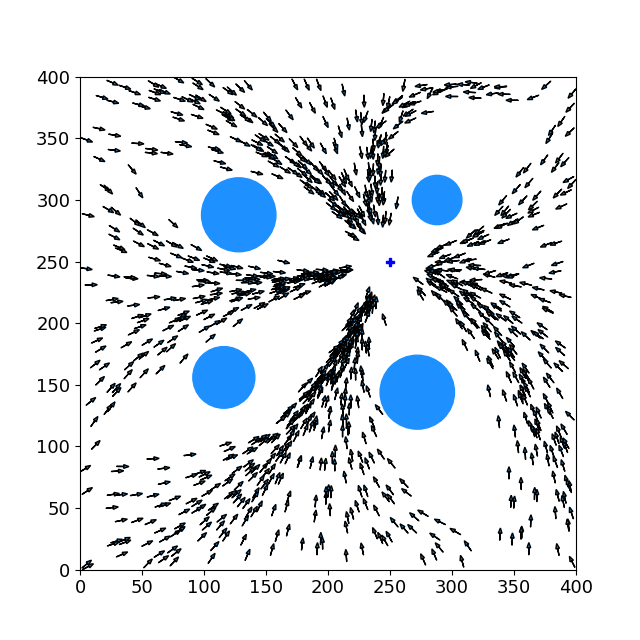}\label{fig:unc_heading0}} 
    \hspace{1cm}
    \subfigure[]{\includegraphics[width=0.6\textwidth]{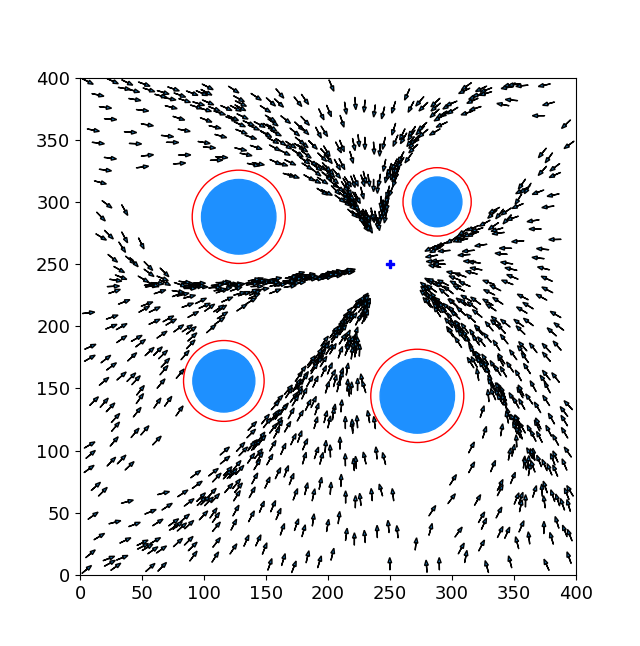}\label{fig:unc_heading1}} 
    \caption{(axis unit: $\times10$ $m$) (a) Results with probabilistic agent's position.    (b) Results with uncertainty added to obstacles. $+$: goal; blue: no-passing area; arrow: the selected heading direction; red: separation requirement  due to uncertainty in UAS operation.}\label{fig:uncheading}
\end{figure}

The testing results are shown in \cref{fig:uncheading}. In \cref{fig:unc_heading0}, the agent's position with uncertainty is plotted. While in \cref{fig:unc_heading1}, the uncertainty of 75 $m$ is added to the obstacle, which is indicated by the red circle. As expected, the UAS tries to keep 75 $m$ away from the obstacles. So either method can work when doing the simulations of collision avoidance with uncertainty. One failure happens near the upper-left obstacle in \cref{fig:unc_heading0}. There are three failures near the lower-left obstacle in \cref{fig:unc_heading1}. The common among the failures is that the agent's origin is approximately on the line passing the obstacle center and the goal. The possible reason is that the policy network gets stuck at the local optimum since the two trajectories next to it behave well.  

\subsection{Moving obstacle avoidance}
\begin{figure}[hbt!]
    \centering
    \subfigure[]{\includegraphics[width=0.65\textwidth]{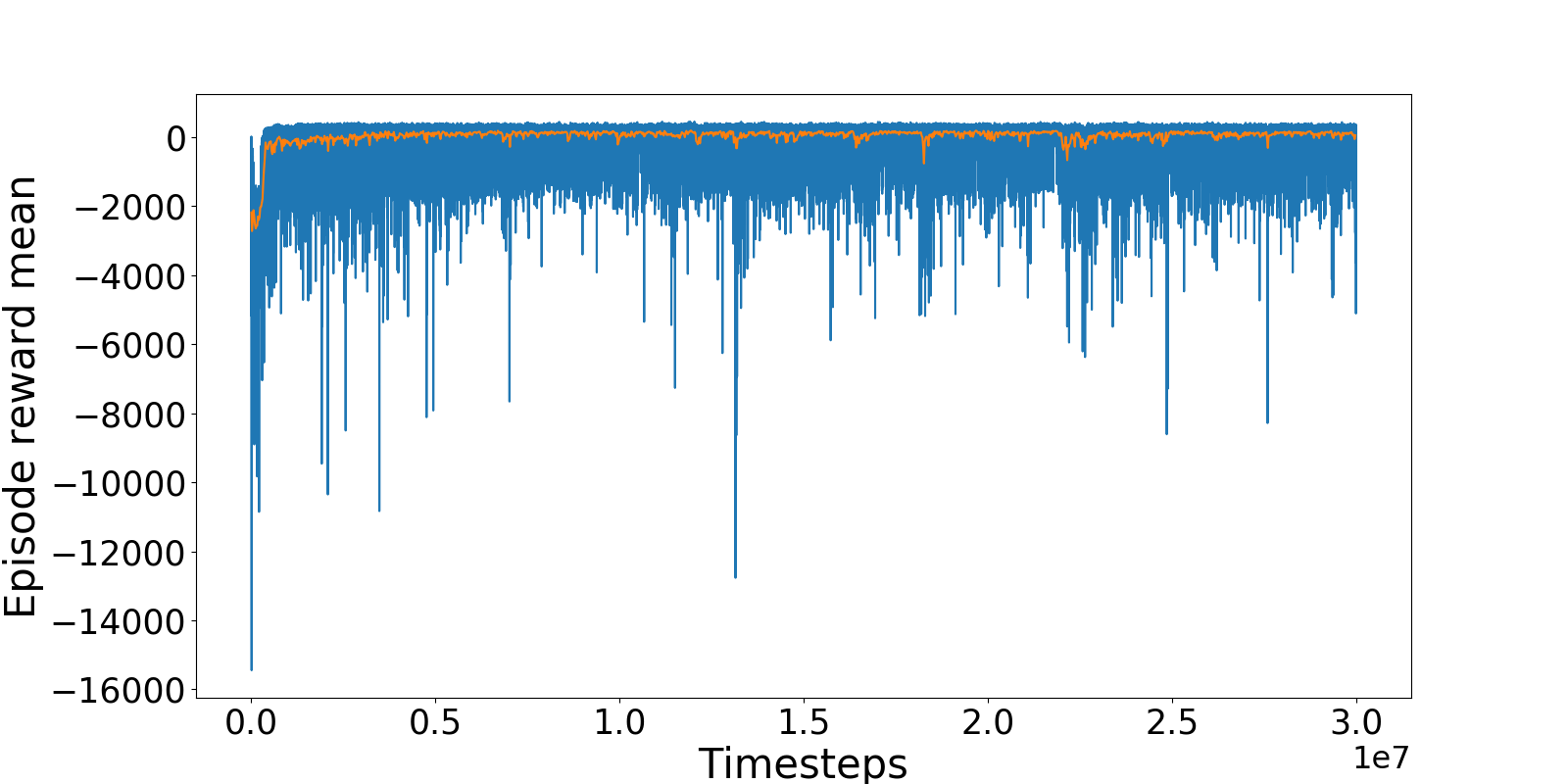}\label{fig:rew_3intruV}} 
    \subfigure[]{\includegraphics[width=0.65\textwidth]{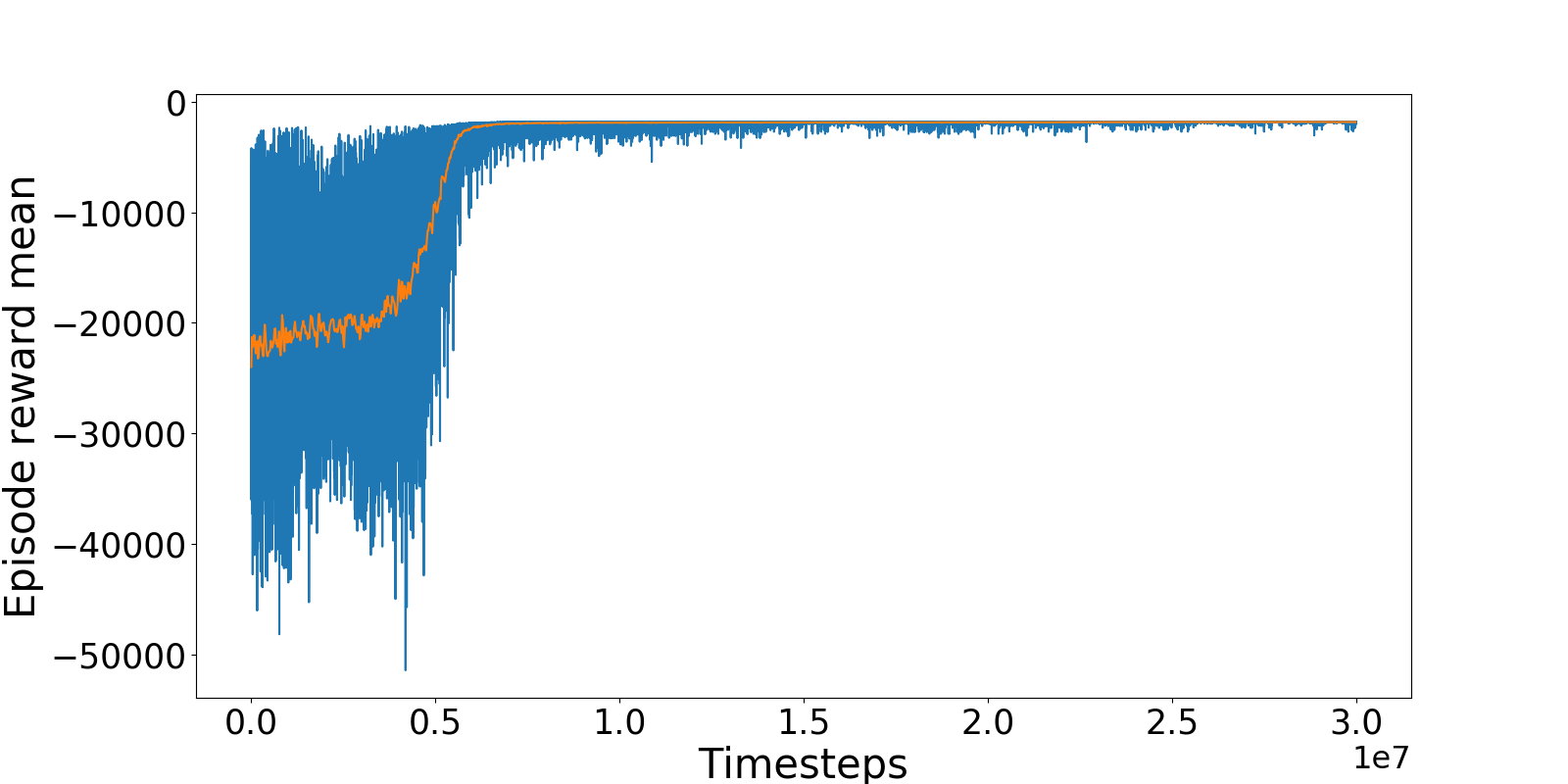}\label{fig:rew_3intruVA}} 
    \caption{(a)Episode reward mean of stochastic-intruder avoidance with control of heading angle. (b) Episode reward mean of deterministic-intruder avoidance with control of heading angle and speed.}\label{fig:movinglc}
\end{figure}

For the moving intruder aircraft avoidance case, the speed of intruders is set to 20 $m/s$. There are two cases for moving obstacle avoidance: stochastic intruder case with control of heading angle and deterministic intruder case with control of heading angle and speed. In the stochastic intruder case, the scenario changes every episode. In detail, the intruder has a different origin and heading angle for each episode. But within one episode, intruders have fixed heading angles.
The reward coefficients are listed in \cref{RC:table}. The episode reward mean is shown in \cref{fig:movinglc}. To visualize the performance of the proposed conflict resolution model, we generate a testing set of 500 episodes following the setting during the training process for each case. Also, the minimum distance of the agent to the three intruders within each episode is collected.   

\begin{table}[hbt!]
    \begin{center}  
    \caption{Reward coefficient}
    \label{RC:table}
        \begin{tabular}{p{5.3cm}p{0.9cm}p{0.83cm}p{0.7cm}p{0.7cm}p{0.7cm}}
            \hline\hline
            \centering
             Coefficient & $c_g$  & $c_0$  & $c_{1i}$ & $c_{2i}$ & $c_{3i}$  \\ \hline 
            \centering
             Stochastic-intruder avoidance & 0.007 & 0.15 & 17 & 0.1 & 12 \\
            \centering
             Deterministic-intruder avoidance & 0.22 & 0.05 & 3 & 0.1 & 12\\
            \hline\hline
        \end{tabular}
    \end{center}
\end{table}

\subsubsection{Stochastic-intruder avoidance with control of heading angle}
\label{three_case}
The origin and heading angle of the three intruders are assumed to follow a uniform distribution and the distribution range is shown in \cref{intruder_information}.  The origin coordinate of agent is uniformly sampled from $(75 \sim 135, 0 \sim 25)$. The goal is located at $(100, 200)$. The agent moves at 20 $m/s$. The intruders are designed to pass the line connecting the UAS origin and the goal.

\begin{table}[hbt!] 
    \begin{center}  
    \caption{Intruder information}
    \label{intruder_information}
        \begin{tabular}{p{3cm}p{2cm}p{2.9cm}p{2.5cm}}
            \hline\hline
            \centering
             Intruder & 1  & 2  & 3  \\ \hline 
            \centering
             Origin coordinate range & $(5 \sim 35, 200)$ & $(20 \sim 65, 20 \sim 65)$ & $(120 \sim 180,$ $120 \sim 180)$ \\
            \centering
             Heading angle range & $[-90^\circ, 0^\circ]$ & $[0^\circ, 90^\circ]$ & $[-180^\circ, -90^\circ]$\\
            \hline\hline
        \end{tabular}
    \end{center}
\end{table}

\begin{figure}[hbt!]
    \centering
    \includegraphics[width=0.9\textwidth]{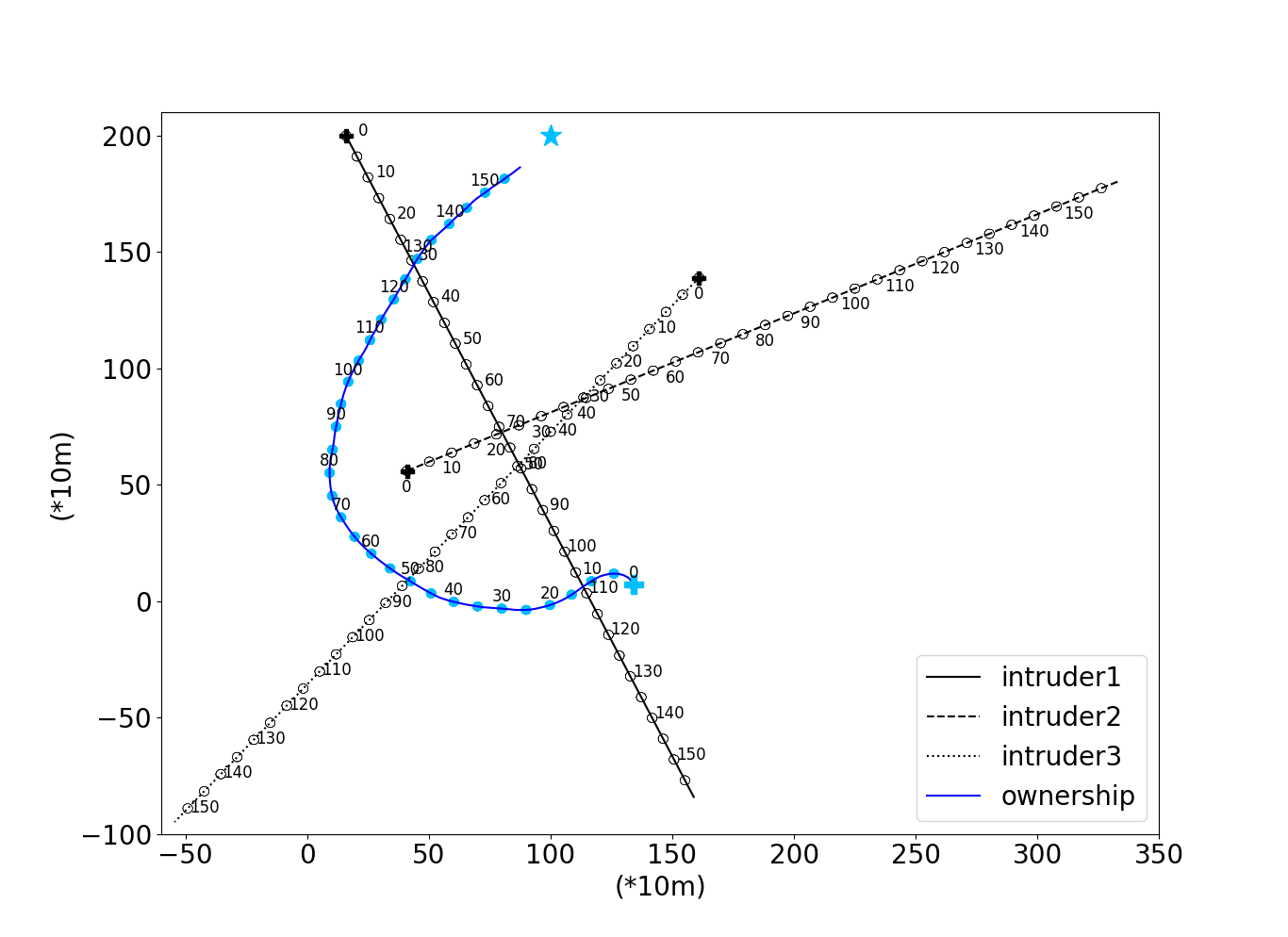}
    \caption{Demonstration of one scenario and UAS performance of stochastic-intruder avoidance with control of heading angle. Number: time step. Blue: ownership; black: intruders. Circle center: UAS position; circle radius: UAS speed. +: origin; *: goal.}
    \label{fig:3V}
\end{figure}

\begin{figure}[hbt!]
    \centering
    \includegraphics[width=0.6\textwidth]{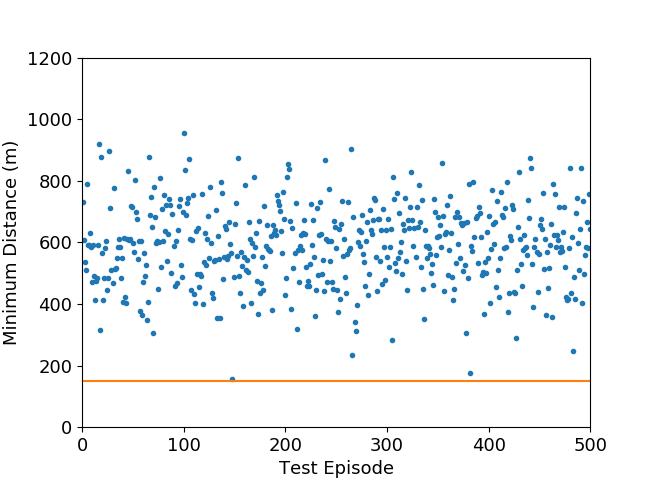}
    \caption{Minimum distance results of stochastic-intruder avoidance with control of heading angle. Orange line: separation requirement of 150 $m$. Blue dot: the minimum distance from the agent to the three intruders within each episode.}
    \label{fig:3Vdis}
\end{figure}

The demonstration of one scenario and UAS performance is shown in \cref{fig:3V}. Information related to the ownership is plotted in blue and black represents the information of intruders. The plus sign denotes the origin and the star sign is the goal for the agent. The centers of circles are the positions of aircraft which are plotted every 5 time steps and labeled with time step every 10 time steps. The radius of the circle represents the aircraft speed. In this scenario, the agent learns to go around the left side to avoid the three intruders. The result of the minimum distance from the agent to the three intruders within each episode is plotted in \cref{fig:3Vdis} by the blue dots. The orange line is the separation requirement of 150 $m$. All the blue dots are above the orange line, which represents that there is no failure case in \cref{fig:3Vdis} and the model succeeds to avoid the three intruders in 500 different testing scenarios.

\subsubsection{Deterministic-intruder avoidance with control of heading angle and speed}
We also investigate the possibility of utilizing the proposed reward function to generate heading angle change command and speed command. This investigation is valuable when changing the heading angle cannot efficiently resolve conflicts. Moreover, with an extra choice of changing speed, the UAS may result in less influence on flight plans of other aircraft and aerospace capacity. However, due to the larger action space, the training process needs more effort.  

The origin and heading angle of the three intruders are listed in \cref{intruderVA_information}. The origin coordinate of agent is $(100, 210)$ and the goal is located at $(100, 0)$. Intruder 1 is designed to test if the ownership can fly at a suitable speed and the other two intruders are set to test the performance of the heading angle change command.

\begin{table}[hbt!] 
    \begin{center}  
    \caption{Intruder information}
    \label{intruderVA_information}
        \begin{tabular}{p{3cm}p{2cm}p{3cm}p{2.6cm}}
            \hline\hline
            \centering
             Intruder & 1  & 2  & 3  \\ \hline 
            \centering
             Origin coordinate & $(90, 170)$ & $(35, 155)$ & $(-15, 115)$ \\
            \centering
             Heading angle & $-90^\circ$ & $-0.2^\circ$ & $-0.2^\circ$\\
            \hline\hline
        \end{tabular}
    \end{center}
\end{table}

\begin{figure}[H]
    \centering
    \includegraphics[width=0.82\textwidth]{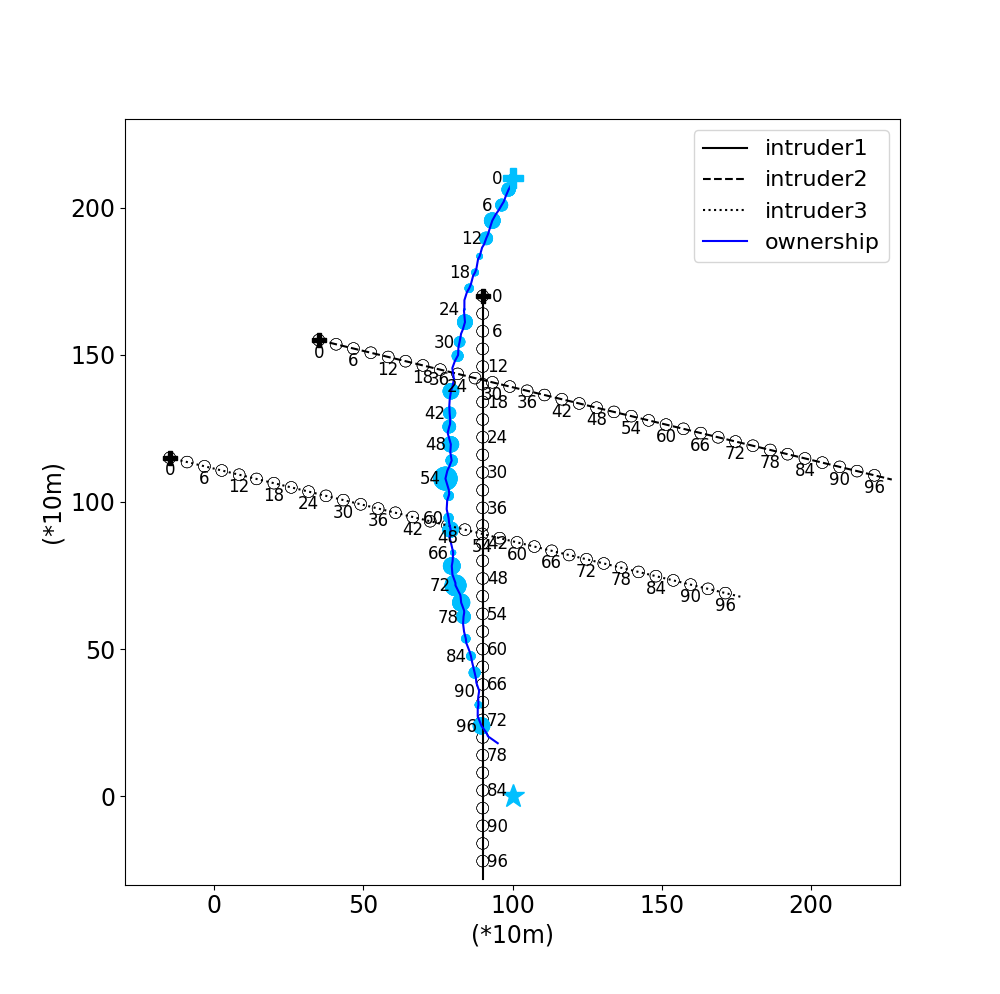}
    \caption{Demonstration of the scenario and UAS performance of deterministic-intruder avoidance with control of heading angle and speed. Number: time step. Blue: ownership; black: intruders. Circle center: UAS position; circle radius: UAS speed. +: origin; *: goal.}
    \label{fig:3VA}
\end{figure}

\begin{figure}[H]
    \centering
    \includegraphics[width=0.6\textwidth]{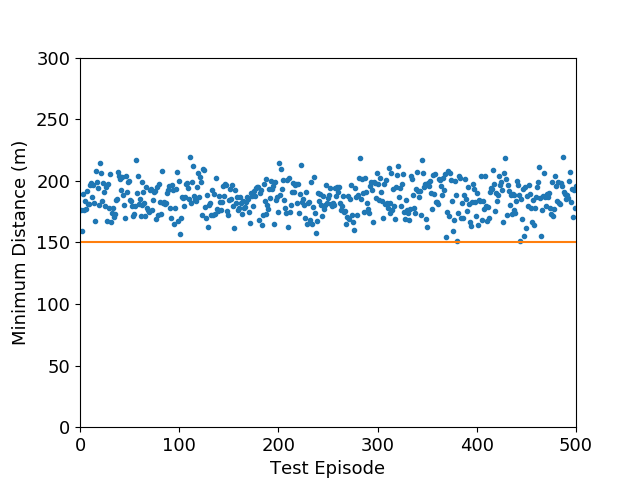}
    \caption{Minimum distance results of deterministic-intruder avoidance with control of heading angle and speed. Orange line: separation requirement of 150 $m$. Blue dot: the minimum distance from the agent to the three intruders within each episode.}
    \label{fig:3VAdis}
\end{figure}

Similar to the result in \cref{fig:3V}, the demonstration of the scenario and UAS performance is shown in \cref{fig:3VA}. Information related to the ownership is plotted in blue and black represents the information of intruders. The plus sign denotes the origin and the star sign is the goal for the agent. The centers of circles are the positions of aircraft which are plotted every 3 time steps and labeled with time step every 6 time steps. The radius of the circle represents the aircraft speed. It can be seen that the agent reduces speed from 12 to 24 time steps to keep a safe separation away from intruder 1. Also, the agent goes around the right side to avoid the approaching intruder 2 and after resolving the possible conflicts with intruder 3 at 66 time step, it flies towards the goal to save time. The result of the minimum distance is plotted in \cref{fig:3VAdis} by the blue dots. All the blue dots are above the orange line which represents the separation requirement of 150 $m$. So, there is no failure case in \cref{fig:3VAdis}, indicating the model succeeds to avoid the three intruders under the control of heading angle and speed.

\section{Conclusion}
\label{Conclusion}
In this work, we present a method for using deep reinforcement learning to allow the UAS to navigate successfully in urban airspace with continuous action space. Both static and moving obstacles are simulated and the trained UAS has the capability to achieve the goal and do conflict resolution simultaneously. We also investigate the performance on different static obstacle shapes and sizes, and under uncertainty in UAS operation. Stochastic intruders are considered in the training process of the moving obstacle experiments. Moreover, we investigate the possibility of the proposed reward function to resolve conflict through heading angle and speed. Results show that the proposed model can provide accurate and robust guidance and resolve conflict with a success rate of over 99\%. To make the proposed algorithm more practical and efficient in the real-world, in future work, we would model part of the intruders as agents and there could be cooperation among the multiple aircraft.

\section*{Acknowledgments}
The research reported in this paper was supported by funds from NASA University Leadership Initiative program (Contract No. NNX17AJ86A, PI: Yongming Liu, Technical Officer: Anupa Bajwa). The support is gratefully acknowledged.

\bibliography{mybibfile}

\end{document}